\documentclass[10pt,twocolumn,letterpaper]{article}

\usepackage{cvpr}
\usepackage{times}
\usepackage{epsfig}
\usepackage{graphicx}
\usepackage{amsmath}
\usepackage{amssymb}
\usepackage{eucal}
\usepackage{subfigure}
\usepackage{multirow}

\usepackage[breaklinks=true,bookmarks=false]{hyperref}

\cvprfinalcopy % *** Uncomment this line for the final submission

\def\cvprPaperID{662} % *** Enter the CVPR Paper ID here

\ifcvprfinal\fi
\begin{document}

%%%%%%%%% TITLE
\title{An Attention Enhanced Graph Convolutional LSTM Network for Skeleton-Based Action Recognition}

\author{Chenyang Si$^{1,2}$ \and Wentao Chen$^{1,3}$ \and Wei Wang$^{1,2}$\thanks{Corresponding Author: Wei Wang} \and Liang Wang$^{1,2}$ \and Tieniu Tan$^{1,2,3}$
    \and
    $^1$Center for Research on Intelligent Perception and Computing (CRIPAC), \\
    National Laboratory of Pattern Recognition (NLPR),\\
    Institute of Automation, Chinese Academy of Sciences (CASIA)\\
    $^2$University of Chinese Academy of Sciences (UCAS)\\
    $^3$University of Science and Technology of China (USTC)\\
    {\tt\small \{chenyang.si, wentao.chen\}@cripac.ia.ac.cn, \{wangwei, wangliang, tnt\}@nlpr.ia.ac.cn}
}

%\author{Chenyang Si$^{1,3}$ \and Wentao Chen$^{1,4}$ \and Wei Wang$^{1,3}$\thanks{Corresponding Author: Wei Wang} \and Liang Wang$^{1,2,3}$ \and Tieniu Tan$^{1,2,3,4}$
%    \and
%    $^1$Center for Research on Intelligent Perception and Computing (CRIPAC), \\
%    National Laboratory of Pattern Recognition (NLPR)\\
%    $^2$Center for Excellence in Brain Science and Intelligence Technology (CEBSIT),\\
%    Institute of Automation, Chinese Academy of Sciences (CASIA)\\
%    $^3$University of Chinese Academy of Sciences (UCAS)\\
%    $^4$University of Science and Technology of China (USTC)\\
%    {\tt\small \{chenyang.si, wentao.chen\}@cripac.ia.ac.cn, \{wangwei, wangliang, tnt\}@nlpr.ia.ac.cn}
%}

%% omit the following lines up until the closing ``}''.
%% Additional authors and addresses can be added with ``\and'',
%% just like the second author.
%% To save space, use either the email address or home page, not both
%\and

\maketitle
\thispagestyle{empty}

%%%%%%%%% ABSTRACT
\begin{abstract}
      Skeleton-based action recognition is an important task that requires the adequate understanding of movement characteristics of a human action from the given skeleton sequence. Recent studies have shown that exploring spatial and temporal features of the skeleton sequence is vital for this task. Nevertheless, how to effectively extract discriminative spatial and temporal features is still a challenging problem. In this paper, we propose a novel Attention Enhanced Graph Convolutional LSTM Network (AGC-LSTM) for human action recognition from skeleton data. The proposed AGC-LSTM can not only capture discriminative features in spatial configuration and temporal dynamics but also explore the co-occurrence relationship between spatial and temporal domains. We also present a temporal hierarchical architecture to increase temporal receptive fields of the top AGC-LSTM layer, which boosts the ability to learn the high-level semantic representation and significantly reduces the computation cost. Furthermore, to select discriminative spatial information, the attention mechanism is employed to enhance information of key joints in each AGC-LSTM layer. Experimental results on two datasets are provided: NTU RGB+D dataset and Northwestern-UCLA dataset. The comparison results demonstrate the effectiveness of our approach and show that our approach outperforms the state-of-the-art methods on both datasets.
\end{abstract}

%%%%%%%%% BODY TEXT
\section{Introduction}
\begin{figure}[!t]
	\centering
	\includegraphics[width=0.9\linewidth]{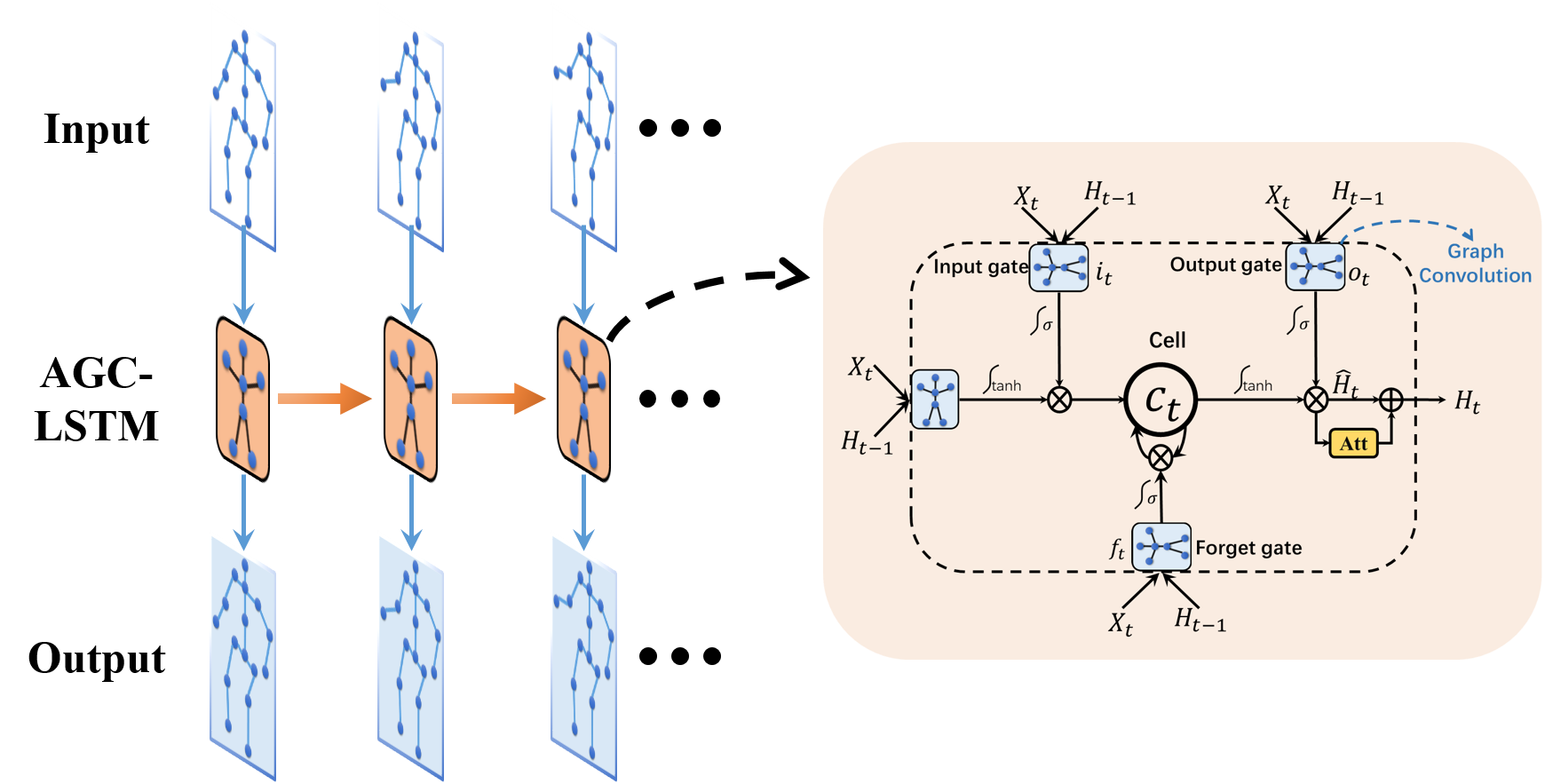}
   	\caption{The structure of one AGC-LSTM layer. Different from traditional LSTM, the graph convolutional operator within AGC-LSTM causes the input, hidden state, and cell memory of AGC-LSTM to be graph-structured data.
   }
   	\label{GCN-LSTM}
\end{figure}

In the computer vision field, human action recognition plays a fundamental and important role, with the purpose of predicting the action classes from videos. It has been studied for decades and is still very popular due to its extensive potential applications, \eg, video surveillance, human-computer interaction, sports analysis and so on \cite{Poppe2010survey, Weinland2011survey, Aggarwal2011Human}.

\begin{figure*}[!t]
	\centering
	\includegraphics[width=0.8\linewidth, height=0.3\linewidth]{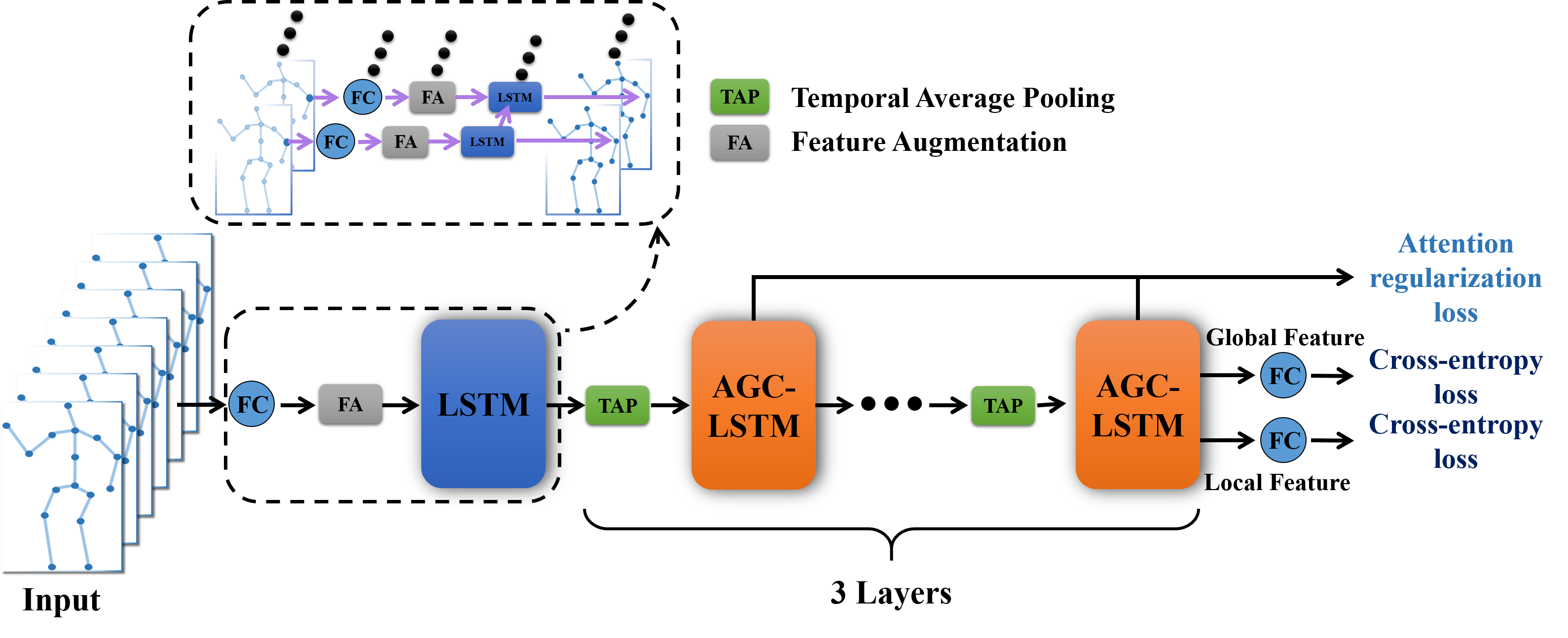}
   	\caption{The architecture of the proposed attention enhanced graph convolutional LSTM network (AGC-LSTM). Feature augmentation (FA) computes feature differences with position features and concatenates both position features and feature differences. LSTM is used to dispel scale variance between feature differences and position features. Three AGC-LSTM layers can model discriminative spatial-temporal features. Temporal average pooling is the implementation of average pooling in the temporal domain. We use the global feature of all joints and the local feature of focused joints from the last AGC-LSTM layer to predict the class of human action.
   }
   	\label{pipeline}
\end{figure*}

Action recognition is a challenging task in the computer vision community. There are various attempts on human action recognition based on RGB video and 3D skeleton data. The RGB video based action recognition methods \cite{Simonyan2014Two-stream, Limin2016Temporal, Tran_2015_ICCV, human2018Pichao} mainly focus on modeling spatial and temporal representations from RGB frames and temporal optical flow. Despite RGB video based methods have achieved promising results, there still exist some limitations, \eg, background clutter, illumination changes, appearance variation, and so on. 3D skeleton data represents the body structure with a set of 3D coordinate positions of key joints. Since skeleton sequence does not contain color information, it is not affected by the limitations of RGB video. Such robust representation allows to model more discriminative temporal characteristics about human actions. Moreover, Johansson \etal~\cite{johansson1973visual} have given an empirical and theoretical basis that key joints can provide highly effective information about human motion. Besides, the Microsoft Kinect \cite{zhang2012microsoft} and advanced human pose estimation algorithms \cite{cao2017realtime} make it easier to gain skeleton data.

For skeleton based action recognition, the existing methods explore different models to learn spatial and temporal features.
%of skeleton sequences.
Song \etal \cite{Song2017Attention} employ a spatial-temporal attention model based on LSTM to select discriminative spatial and temporal features. The Convolutional Neural Networks (CNNs) are used to learn spatial-temporal features from skeletons in \cite{Yong2015Skeleton, Chao2018Co-occurrence, Ke2017A}. \cite{Yan2018Spatial, Kalpit2018Part} employ graph convolutional networks (GCN) for action recognition.
%\cite{Yan2018Spatial, Kalpit2018Part} employ GCN-based models for action recognition.
%Yan \etal~ \cite{Yan2018Spatial} propose a spatial-temporal graph convolutional network (ST-GCN) for action recognition.
%Each ST-GCN layer applies the graph convolutional network to model spatial configuration of joints and then uses a convolutional operator to learn temporal dynamics.
Compared with \cite{Yan2018Spatial, Kalpit2018Part}, Si \etal~\cite{Chenyang2018Skeleton} propose to utilize the graph neural network and LSTM to represent spatial and temporal information, respectively.
%Despite the significant performance improvement in \cite{Chenyang2018Skeleton}, it ignores the co-occurrence relationship between spatial and temporal domains.
In short, all these methods are trying to design an effective model that can identify spatial and temporal features of skeleton sequence. Nevertheless, how to effectively extract discriminative spatial and temporal features is still a challenging problem.

Generally, there are three notable characteristics for human skeleton sequences: 1) There are strong correlations between each node and its adjacent nodes so that the skeleton frames contain abundant body structural information. 2) Temporal continuity exists not only in the same joints (\eg, hand, wrist and elbow), but also in the body structure. 3) There is a co-occurrence relationship between spatial and temporal domains. In this paper, we propose a novel and general framework called attention enhanced graph convolutional LSTM network (AGC-LSTM) for skeleton-based action recognition, which improves the skeleton representation by synchronously learning spatiotemporal characteristics mentioned above.

The architecture of the proposed AGC-LSTM network is shown in Fig.\ref{pipeline}. Firstly, the coordinate of each joint is transformed into a spatial feature with a linear layer. Then we concatenate spatial feature and feature difference between two consecutive frames to compose an augmented feature. In order to dispel scale variance between both features, a shared LSTM is adopted to process each joint sequence. Next, we apply three AGC-LSTM layers to model spatial-temporal features. As shown in Fig.\ref{GCN-LSTM}, due to the graph convolutional operator within AGC-LSTM, it can not only effectively capture discriminative features in spatial configuration and temporal dynamics but also explore the co-occurrence relationship between spatial and temporal domains. More specially, the attention mechanism is employed to enhance the features of key nodes at each time step, which can promote AGC-LSTM to learn more discriminative features. For example, the features of ``elbow'', ``wrist'' and ``hand'' are very important for action ``handshaking'' and should be enhanced in the process of identifying the behavior. Inspired by spatial pooling in CNNs, we present a temporal hierarchical architecture with temporal average pooling to increase temporal receptive fields of the top AGC-LSTM layers, which boosts the ability to learn high-level spatiotemporal semantic features and significantly reduces the computational cost. Finally, we use the global feature of all joints and the local feature of focused joints from the last AGC-LSTM layer to predict the class of human actions. Although the joint-based model achieves the state-of-the-art results, we also explore the performance of the proposed model on the part level. For the part-based model, the concatenation of joints of each part serves as a node to construct the graph. Furthermore, the two-stream model based on joint and part can lead to further performance improvement.

The main contributions of this work are summarized as follows:
\begin{itemize}
  \item We propose a novel and general AGC-LSTM network for skeleton-based action recognition, which is the first attempt of graph convolutional LSTM for this task.
  \item The proposed AGC-LSTM is able to effectively capture discriminative spatiotemporal features. More specially, the attention mechanism is employed to enhance the features of key nodes, which assists in improving spatiotemporal expressions.
  \item A temporal hierarchical architecture is proposed to boost the ability to learn high-level spatiotemporal semantic features and significantly reduce the computational cost.
  \item The proposed model achieves the state-of-the-art results on both NTU RGB+D dataset and Northwestern-UCLA dataset. We perform extensive experiments to demonstrate the effectiveness of our model.
\end{itemize}

\section{Related Work}

%In this section, we briefly review approaches related to the proposed attention enhanced graph convolutional LSTM network.

\textbf{\emph{Neural networks with graph}} \hspace{3mm}
Recently, graph-based models have attracted a lot of attention due to the effective representation for the graph structure data \cite{Keyulu2018How}. Existing graph models mainly fall into two architectures. One framework called graph neural network (GNN) is the combination of graph and recurrent neural network. Through multiple iterations of message passing and states updating of nodes, each node captures the semantic relation and structural information within its neighbor nodes. Qi \etal~\cite{Qi_2018_ECCV} apply GNN to address the task of detecting and recognizing human-object interactions in images and videos. Li \etal~\cite{Li_2017_ICCV} exploit the GNNs to model dependencies between roles and predict a consistent structured output for situation recognition. The other framework is graph convolutional network (GCN) that generalizes convolutional neural networks to graph. There are two types of GCNs: spectral GCNs and spatial GCNs. Spectral GCNs transform graph signals on graph spectral domains and then apply spectral filters on spectral domains. For example, the CNNs are utilized in the spectral domain relying on the graph Laplacian \cite{NIPS2015_5954, henaff2015deep}. Kipf \etal~\cite{Thomas2017Semi} introduce Spectral GCNs for semi-supervised classification on graph-structured data. For spatial GCNs, the convolution operation is applied to compute a new feature vector for each node using its neighborhood information. Simonovsky \etal~\cite{Simonovsky_2017_CVPR} formulate a convolution-like operation on graph signals performed in the spatial domain and are the first to apply graph convolutions to point cloud classification. In order to capture the spatial-temporal features of graph sequences, a graph convolutional LSTM is firstly proposed in \cite{Youngjoo2016Structured}, which is an extension of GCNs to have the recurrent architecture. Inspired by \cite{Youngjoo2016Structured}, we exploit a novel AGC-LSTM network to learn inherent spatiotemporal representations from skeleton sequences.

\textbf{\emph{Skeleton-based action recognition}} \hspace{3mm}
Human action recognition based on skeleton data has received a lot of attention, due to its effective representation of motion dynamics. Traditional skeleton-based action recognition methods mainly focus on designing hand-crafted features \cite{Raviteja2014Human, Wang2012Mining, Hussein2013Human}. Vemulapalli \etal~\cite{Raviteja2016Rolling} represent each skeleton using the relative 3D rotations between various body parts. The relative 3D geometry between all pairs of body parts is applied to represent the 3D human skeleton in \cite{Raviteja2014Human}. %Chaudhry\etal~\cite{Rizwan2013inspired} learn the dynamics of skeletons using a set of Linear Dynamical Systems.

Recent works mainly learn human action representations with deep learning networks\cite{Action2018Zhengyuan, Chunyu2018Memory, Human2017Fabien}. Du \etal~\cite{Du2015Hierarchical} divide human skeleton into five parts according to the human physical structure, and then separately feed them into a hierarchical recurrent neural network to recognize actions. A spatial-temporal attention network learns to selectively focus on discriminative spatial and temporal features in \cite{Song2017Attention}. Zhang \etal~\cite{Zhang2017View} present a view adaptive model for skeleton sequence, which is capable of regulating the observation viewpoints to the suitable ones by itself. The works in \cite{Yan2018Spatial, Kalpit2018Part, Chao2018Co-occurrence, Chenyang2018Skeleton} further show that learning discriminative spatial and temporal features is the key element for human action recognition. A hierarchical CNN model is presented in \cite{Chao2018Co-occurrence} to learn representations for joint co-occurrences and temporal evolutions. A spatial-temporal graph convolutional network (ST-GCN) is proposed for action recognition in \cite{Yan2018Spatial}. Each spatial-temporal graph convolutional layer constructs spatial characteristics with a graph convolutional operator, and models temporal dynamic with a convolutional operator. In addition, a part-based graph convolutional network (PB-GCN) is proposed to learn the relations between parts in \cite{Kalpit2018Part}. Compared with ST-GCN \cite{Yan2018Spatial} and PB-GCN \cite{Kalpit2018Part}, Si \etal~\cite{Chenyang2018Skeleton} apply graph neural networks to capture spatial structural information and then use LSTM to model temporal dynamics. Despite the significant performance improvement in \cite{Chenyang2018Skeleton}, it ignores the co-occurrence relationship between spatial and temporal features. In this paper, we propose a novel attention enhanced graph convolutional LSTM network that can not only effectively extract discriminative spatial and temporal features but also explore the co-occurrence relationship between spatial and temporal domains.

\section{Model Architecture}

%In this section, we first briefly review the graph convolutional neural network and then introduce our attention enhanced graph convolutional LSTM. Finally, we illustrate the architecture of the proposed AGC-LSTM network.

\subsection{Graph Convolutional Neural Network}

Graph convolutional neural network (GCN) is a general and effective framework for learning representation of graph structured data. Various GCN variants have achieved the state-of-the-art results on many tasks. For skeleton-based action recognition, let $\mathcal{G}_t$ = \{$\mathcal{V}_t, \mathcal{E}_t$\} denotes a graph of human skeleton on a single frame at time $t$, where $\mathcal{V}_t$ is the set of $N$ joint nodes and $\mathcal{E}_t$ is the set of skeleton edges. The neighbor set of a node $v_{ti}$ is defined as $\mathcal{N}(v_{ti}) = \{v_{tj}|d(v_{ti}, v_{tj}) \leq D\}$, where $d(v_{ti}, v_{tj})$ is the minimum path length from $v_{tj}$ to $v_{ti}$. A graph labeling function $\ell: \mathcal{V}_t \to \{1,2,...,K\}$ is designed to assign the labels $\{1,2,...,K\}$ to each graph node $v_{ti} \in \mathcal{V}_t$, which can partition the neighbor set $\mathcal{N}(v_{ti})$ of node $v_{ti}$ into a fixed number of $K$ subsets. The graph convolution is generally computed as:
\begin{align}
\label{GCN1}
  \textbf{Y}_{out}(v_{ti}) = \sum_{v_{tj}\in\mathcal{N}(v_{ti})} \frac{1}{Z_{ti}(v_{tj})} \textbf{X}(v_{tj}) \textbf{W}(\ell(v_{tj}))
\end{align}
where $\textbf{X}(v_{tj})$ is the feature of node $v_{tj}$. $\textbf{W}(\cdot)$ is a weight function that allocates a weight indexed by the label $\ell(v_{tj})$ from $K$ weights. $Z_{ti}(v_{tj})$ is the number of the corresponding subset, which normalizes feature representations. $\textbf{Y}_{out}(v_{ti})$ denotes the output of graph convolution at node $v_{ti}$. More specifically, with the adjacency matrix, the Eqn. \ref{GCN1} can be represented as:
\begin{align}
\label{GCN2}
  \textbf{Y}_{out} = \sum_{k=1}^{K} \boldsymbol{\Lambda}_k^{-\frac{1}{2}} \textbf{A}_k \boldsymbol{\Lambda}_k^{-\frac{1}{2}} \textbf{X} \textbf{W}_k
\end{align}
where $\textbf{A}_k$ is the adjacency matrix in spatial configuration of the label $k \in \{1,2,...,K\}$. $\boldsymbol{\Lambda}_k^{ii} = \sum_j \textbf{A}_k^{ij}$ is a degree matrix.

\subsection{Attention Enhanced Graph Convolutional LSTM}

For sequence modeling, a lot of studies have demonstrated that LSTM, as a variant of RNN, has an amazing ability to model long-term temporal dependencies. Various LSTM-based models are employed to learn temporal dynamics of skeleton sequences. However, due to the fully connected operator within LSTM, there is a limitation of ignoring spatial correlation for skeleton-based action recognition. Compared with LSTM, AGC-LSTM can not only capture discriminative features in spatial configuration and temporal dynamics, but also explore the co-occurrence relationship between spatial and temporal domains.

\begin{figure}[t]
	\centering
	\includegraphics[width=0.9\linewidth, height=0.5\linewidth]{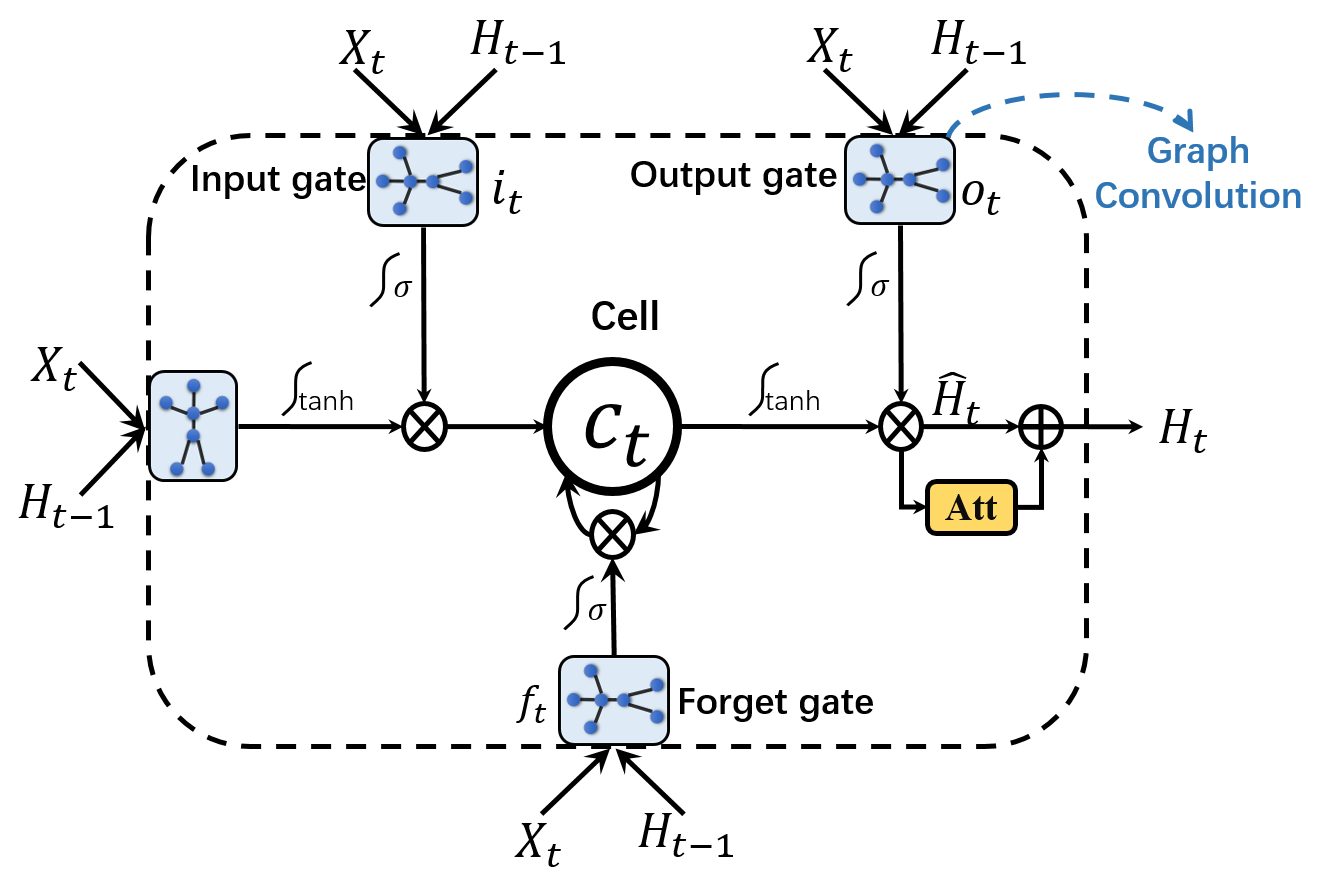}
   	\caption{The structures of AGC-LSTM unit. Compared with LSTM, the inner operator of AGC-LSTM is graph convolutional calculation. To highlight more discriminative information, the attention mechanism is employed to enhance the features of key nodes.
   }
   	\label{GCN-LSTM-cell}
\end{figure}

Like LSTM, AGC-LSTM also contains three gates: an input gate $\textbf{i}_t$, a forget gate $\textbf{f}_t$, an output gate $\textbf{o}_t$. However, these gates are obtained with the graph convolution operator. The input $\textbf{X}_t$, hidden state $\textbf{H}_t$, and cell memory $\textbf{C}_t$ of AGC-LSTM are graph-structured data. Fig.\ref{GCN-LSTM-cell} shows the strcture of AGC-LSTM unit. Due to the graph convolutional operator within AGC-LSTM, the cell memory $\textbf{C}_t $ and hidden state $\textbf{H}_t$ are able to exhibit temporal dynamics, as well as contain spatial structural information. The functions of AGC-LSTM unit are defined as follows:

%not only exhibit temporal dynamics but also contain spatial structural information. The functions of AGC-LSTM unit are defined as follows:

\begin{align}
\label{gc-lstm_neuron}
  \textbf{i}_t & = \sigma(\textbf{W}_{xi} *_{\mathcal{G}} \textbf{X}_t + \textbf{W}_{hi} *_{\mathcal{G}} \textbf{H}_{t-1} + \textbf{b}_i) \nonumber\\
  \textbf{f}_t & = \sigma(\textbf{W}_{xf} *_{\mathcal{G}} \textbf{X}_t + \textbf{W}_{hf} *_{\mathcal{G}} \textbf{H}_{t-1} + \textbf{b}_f) \nonumber\\
  \textbf{o}_t & = \sigma(\textbf{W}_{xo} *_{\mathcal{G}} \textbf{X}_t + \textbf{W}_{ho} *_{\mathcal{G}} \textbf{H}_{t-1} + \textbf{b}_o) \nonumber\\
  \textbf{u}_t & = tanh(\textbf{W}_{xc} *_{\mathcal{G}} \textbf{X}_t + \textbf{W}_{hc} *_{\mathcal{G}} \textbf{H}_{t-1} + \textbf{b}_c) \\
  \textbf{C}_t & = \textbf{f}_t \odot \textbf{C}_{t-1} + \textbf{i}_t \odot \textbf{u}_t  \nonumber\\
  \widehat{\textbf{H}}_t & = \textbf{o}_t \odot tanh(\textbf{C}_t) \nonumber\\
  \textbf{H}_t & = f_{att} \left( \widehat{\textbf{H}}_t \right) + \widehat{\textbf{H}}_t \nonumber
\end{align}
where $*_{\mathcal{G}}$ denotes the graph convolution operator and $\odot$ denotes the Hadamard product. $\sigma \left( \cdot \right)$ is the sigmoid activation function. $\textbf{u}_t$ is the modulated input. $\widehat{\textbf{H}}_t$ is an intermediate hidden state. $\textbf{W}_{xi} *_{\mathcal{G}} \textbf{X}_t$  denotes a graph convolution of $\textbf{X}_t $ with $\textbf{W}_{xi} $, which can be written as Eqn.\ref{GCN1}. $f_{att}(\cdot)$ is an attention network that can select discriminative information of key nodes. The sum of $f_{att} \left( \widehat{\textbf{H}}_t \right)$ and $\widehat{\textbf{H}}_t$ as the output aims to strengthen information of key nodes without weakening information of non-focused nodes, which can maintain the integrity of spatial information.

\begin{figure}[t]
	\centering
	\includegraphics[width=0.8\linewidth]{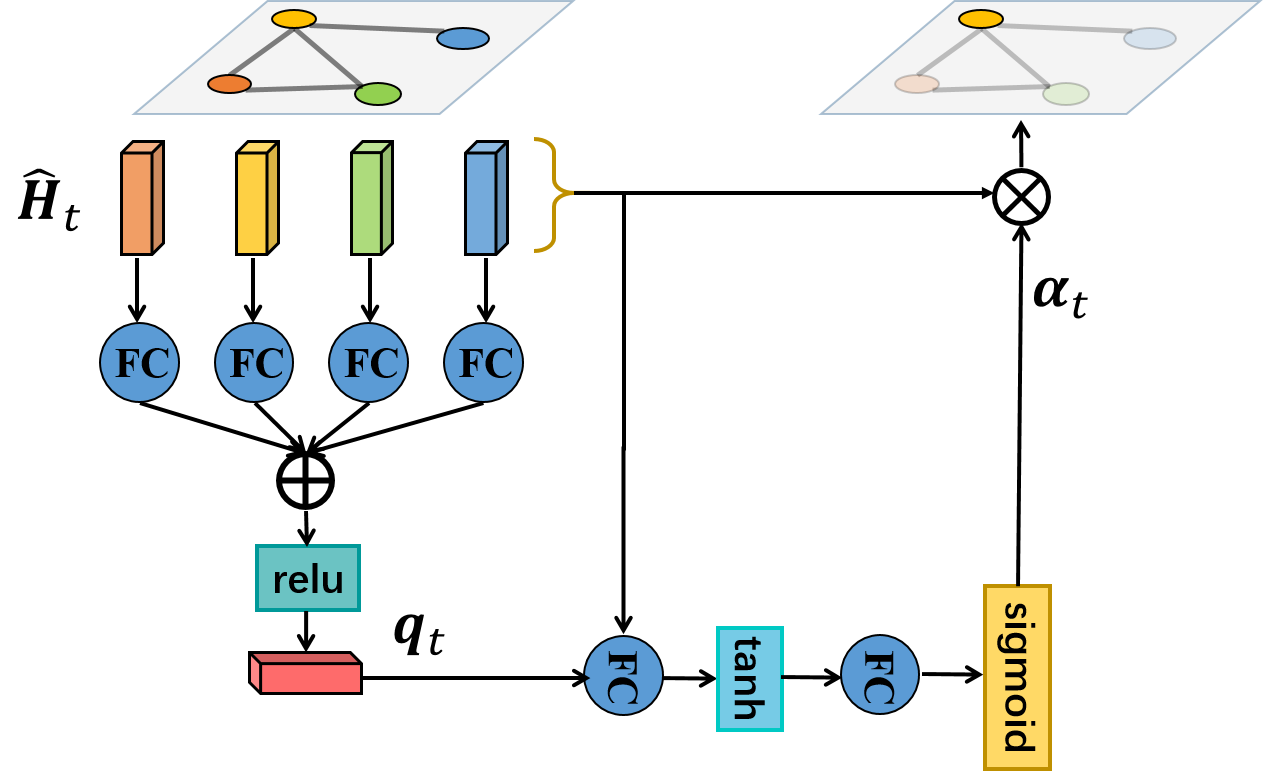}
   	\caption{ Illustration of the spatial attention network.}
   	\label{attention}
\end{figure}

The attention network is employed to adaptively focus on key joints with a soft attention mechanism that can automatically measure the importance of joints. The illustration of the spatial attention network is shown in Fig.\ref{attention}. The intermediate hidden state $\widehat{\textbf{H}}_t$ of AGC-LSTM contains rich spatial structural information and temporal dynamics that are beneficial in guiding the selection of key joints. So we first aggregate the information of all nodes as a query feature:
\begin{align}
\label{aggregate_node}
    \textbf{q}_t & = ReLU \left( \sum_{i=1}^{N}\textbf{W}\widehat{\textbf{H}}_{ti} \right)
\end{align}
where $\textbf{W}$ is the learnable parameter matrix. Then the attention scores of all nodes can be calculated as:
\begin{align}
\label{alpha_compute}
    \boldsymbol{\alpha}_t = Sigmoid \left(\textbf{U}_s tanh \left(\textbf{W}_h \widehat{\textbf{H}}_t + \textbf{W}_q \textbf{q}_t + \textbf{b}_s \right) + \textbf{b}_u \right)
\end{align}
where $\boldsymbol{\alpha}_t = (\alpha_{t1}, \alpha_{t2}, ..., \alpha_{tN})$, and $\textbf{U}_s, \textbf{W}_h, \textbf{W}_q$ are the learnable parameter matrixes. $\textbf{b}_s, \textbf{b}_u$ are the bias. We use the non-linear function of \emph{Sigmoid} due to the possibility of existing multiple key joints. The hidden state $\textbf{H}_{ti}$ of node $v_{ti}$ can also be represented as $(1+\alpha_{ti}) \cdot \widehat{\textbf{H}}_{ti}$. The attention enhanced hidden state $\textbf{H}_t$ will be fed into the next AGC-LSTM layer. Note that, at the last AGC-LSTM layer, the aggregation of all node features will serve as a global feature $\textbf{F}_t^{g}$, and the weighted sum of focused nodes will serve as a local feature $\textbf{F}_t^{l}$:
\begin{align}
    \label{global_feature}
    \textbf{F}_t^{g} & = \sum_{i=1}^{N} \textbf{H}_{ti} \\
    \textbf{F}_t^{l} & = \sum_{i=1}^{N} \alpha_{ti} \cdot \widehat{\textbf{H}}_{ti}
    \label{local_feature}
\end{align}
The global feature $\textbf{F}_t^{g}$ and local feature $\textbf{F}_t^{l}$ are used to predict the class of human action.

\subsection{AGC-LSTM Network}

We propose an end-to-end attention enhanced graph convolutional LSTM network (AGC-LSTM) for skeleton-based human action recognition. The overall pipeline of our model is shown in Fig.\ref{pipeline}. In the following, we discuss the rationale behind the proposed framework in detail.

\textbf{Joints Feature Representation.} For the skeleton sequence, we first map the 3D coordinate of each joint into a high-dimensional feature space using a linear layer and an LSTM layer. The first linear layer encodes the coordinates of joints into a 256-dim vector as position features $\textbf{P}_t \in \mathbb{R}^{N \times 256}$ , and $\textbf{P}_{ti} \in \mathbb{R}^{1 \times 256}$ denotes the position representation of joint $i$. Due to only containing position information, the position feature $\textbf{P}_{ti}$ is beneficial for learning spatially structured characteristic in the graph model. Frame difference features $\textbf{V}_{ti}$ between two consecutive frames can facilitate the acquisition of dynamic information for AGC-LSTM. In order to take into account both advantages, the concatenation of both features serve as an augmented feature to enrich feature information. However, the concatenation of position feature $\textbf{P}_{ti}$ and frame difference feature $\textbf{V}_{ti}$ exists the scale variance of the features vectors. Therefore, we adopt an LSTM layer to dispel scale variance between both features:

\begin{align}
\label{feature_flatten}
    \textbf{E}_{ti} & = f_{lstm} \left( concat \left(\textbf{P}_{ti}, \textbf{V}_{ti} \right) \right) \nonumber\\
    & = f_{lstm} \left( concat \left(\textbf{P}_{ti}, \left( \textbf{P}_{ti} - \textbf{P}_{(t-1)i} \right) \right) \right)
\end{align}
where $\textbf{E}_{ti}$ is the augmented feature of joint $i$ at time $t$. Note that the linear layer and LSTM are shared among different joints.

\textbf{Temporal Hierarchical Architecture.} After the LSTM layer, the sequence $\{\textbf{E}_1,\textbf{E}_2,...,\textbf{E}_{T}\}$ of augmented features will be fed into the following GC-LSTM layers as the node features, where $\textbf{E}_t \in \mathbb{R}^{N \times d_e}$. The proposed model stacks three AGC-LSTM layers to learn the spatial configuration and temporal dynamics. Inspired by spatial pooling in CNNs, we present a temporal hierarchical architecture of AGC-LSTM with average pooling in temporal domain to increase the temporal receptive field of the top AGC-LSTM layers. Through the temporal hierarchical architecture, the temporal receptive field of each time input at the top AGC-LSTM layer becomes a short-term clip from a frame, which can be more sensitive to the perception of the temporal dynamics. In addition, it can significantly reduce computational cost on the premise of improving performance.

\textbf{Learning AGC-LSTM.} Finally, the global feature $\textbf{F}_t^g$ and local feature $\textbf{F}_t^l$ of each time step are transformed into the scores $\textbf{o}_t^g$ and $\textbf{o}_t^l$ for $C$ classes, where $\textbf{o}_t = (o_{t1}, o_{t2},...,o_{tC})$. And the predicted probability being the $i^{th}$ class is then obtained as:
\begin{align}
\label{score_compute}
   {\hat y}_{ti} & = { {e^{o_{ti}}} \over { \sum_{j=1}^C e^{o_{tj}} }}, i = 1,...,C
\end{align}
During training, considering that the hidden state of each time step on the top AGC-LSTM contains a short-term dynamics, we supervise our model with the following loss:
\begin{small}
\begin{align}
\label{loss}
   \displaystyle \mathcal{L} & =  - \sum_{t=1}^{T_3} \sum_{i=1}^{C} y_i log {\hat y}_{ti}^g - \sum_{t=1}^{T_3} \sum_{i=1}^{C} y_i log {\hat y}_{ti}^l \\
   \displaystyle  & + \lambda \sum_{j=1}^{3} \sum_{n=1}^{N} {\left( 1 - {{\sum_{t=1}^{T_j} \alpha_{tnj}} \over {T_j}} \right)}^{2} + \beta \sum_{j=1}^{3} {1 \over {T_j}} \sum_{t=1}^{T_j} {\left( \sum_{n=1}^{N} \alpha_{tnj} \right)}^{2} \nonumber
\end{align}
\end{small}
where $\emph{\textbf{y}} = \left( y_1,...,y_C \right)$ is the groundtruth label. $T_j$ denotes the number of time step on $j^{th}$ AGC-LSTM layer. The third term aims to pay equal attention to different joints. The last term is to limit the number of interested nodes. $\lambda$ and $\beta$ are weight decaying coefficients. Note that only the sum probability of ${\hat{\emph{\textbf{y}}}}_{T_3}^g$ and ${\hat{\emph{\textbf{y}}}}_{T_3}^l$ at the last time step is used to predict the class of the human action.

\begin{figure}[!t]
	\centering
	\includegraphics[width=0.8\linewidth, height=0.4\linewidth]{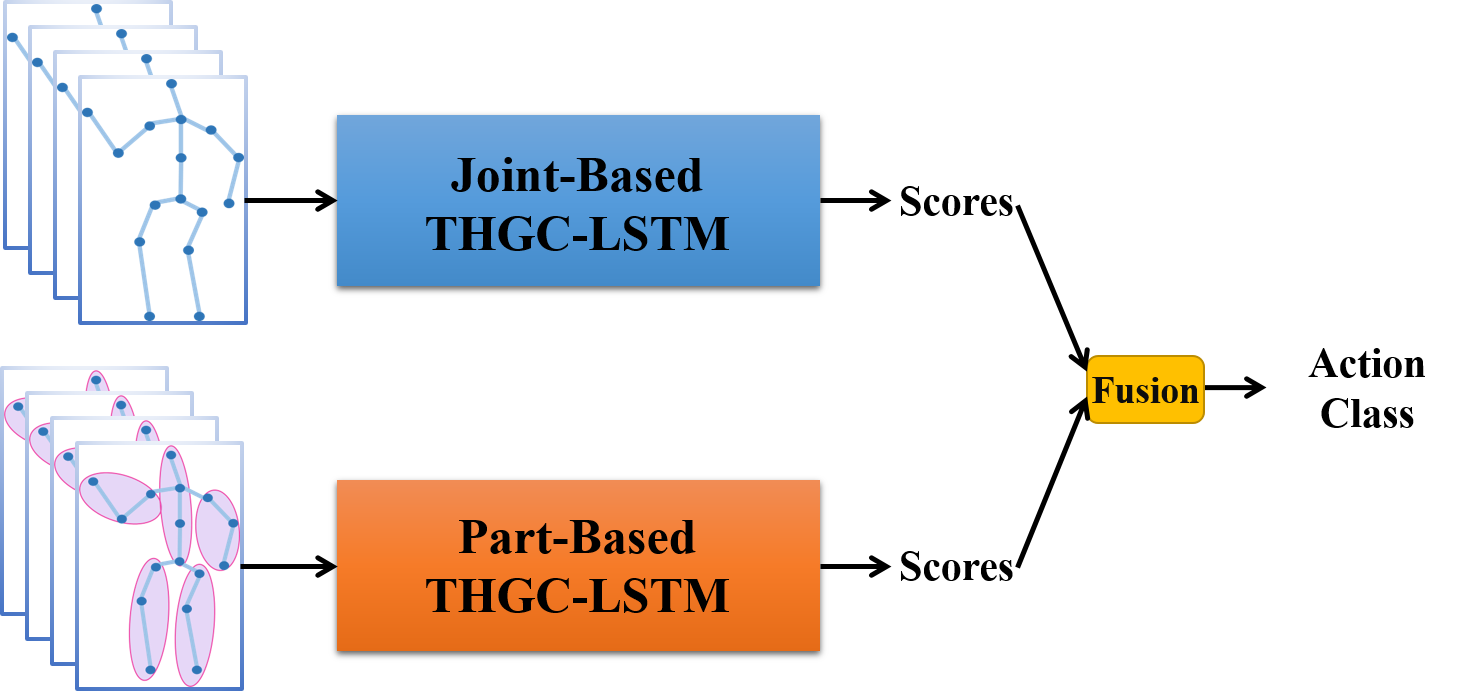}
   	\caption{ Illustration of the hybrid model based on joins and parts.}
   	\label{two-stream}
\end{figure}

Although the joint-based AGC-LSTM network has achieved the state-of-the-art results, we also explore the performance of the proposed model on the part level. According to human physical structure, the body can be divided into several parts. Similar to joint-based AGC-LSTM network, we first capture part features with a linear layer and a shared LSTM layer. Then the part features as node representations are fed into three AGC-LSTM layers to model spatial-temporal characteristics. The results illustrate that our model can also achieve superior performance on the part level. Furthermore, the hybrid model (shown in Fig.\ref{two-stream}) based on joints and parts can lead to further performance improvement.

\section{Experiments}

%We have evaluated our proposed model on two datasets: NTU RGB+D dataset \cite{Shahroudy2016NTU} and Northwestern-UCLA dataset \cite{Jiang2014Cross}. The analysis of experimental results confirms the effectiveness of our model for skeleton-based action recognition.

\subsection{Datasets}

\textbf{NTU RGB+D dataset} \cite{Shahroudy2016NTU}.
This dataset contains 60 different human action classes that are divided into three major groups: daily actions, mutual actions, and health-related actions. There are 56,880 action samples in total which are performed by 40 distinct subjects. Each action sample contains RGB video, depth map sequence, 3D skeleton data, and infrared video captured by three Microsoft Kinect v2 cameras concurrently. The 3D skeleton data that we focus on consists of 3D positions of 25 body joints per frame. There are two evaluation protocols for this dataset: Cross-Subject (CS) and Cross-View (CV) \cite{Shahroudy2016NTU}. Under the Cross-Subject protocol, actions performed by 20 subjects constitute the training set and the rest of actions performed by the other 20 subjects are used for testing. For Cross-View evaluation, samples captured by the first two cameras are used for training and the rest are for testing.

\textbf{Northwestern-UCLA dataset} \cite{Jiang2014Cross}.
This dataset contains 1494 video clips covering 10 categories. It is captured by three Kinect cameras simultaneously from a variety of viewpoints. Each action sample contains RGBD and human skeleton data performed by 10 different subjects. The evaluation protocol is the same as in \cite{Jiang2014Cross}. Samples from the first two cameras constitute the training set and samples from the other camera constitute the testing dataset.

\subsection{Implementation Details}
In our experiments, we sample a fixed length $T$ from each skeleton sequence as the input. We set the length $T = 100$ and 50 for NTU dataset and Northwestern-UCLA dataset, respectively. In the proposed AGC-LSTM, the neighbor set of each node contains only nodes directly connected with itself, so $D = 1$. In order to compare fairly with ST-GCN \cite{Yan2018Spatial}, the graph labeling function in AGC-LSTM will partition the neighbor set into $K = 3$ subsets according to \cite{Yan2018Spatial}: the root node itself, centripetal group, and centrifugal group. The channels of three AGC-LSTM layers are set to 512. During training, we use the Adam optimizer \cite{kingma2015adam} to optimize the network. Dropout with a probability of 0.5 is adopted to avoid over-fitting on these two datasets. We set $\lambda$ and $\beta$ to 0.01 and 0.001, respectively. The initial learning rate is set to 0.0005 and reduced by multiplying it by 0.1 every 20 epochs. The batch sizes for the NTU dataset and Northwestern-UCLA dataset are 64 and 30, respectively.

\subsection{Results and Comparisons}

In this section, we compare our proposed attention enhanced graph convolutional LSTM network (AGC-LSTM) with several state-of-the-art methods on the used two datasets.

\subsubsection{NTU RGB+D Dataset}

\begin{table}[b]
\begin{center}
\begin{tabular}{lccc}
\hline\noalign{\smallskip}
Methods&Year&CV&CS\\
\noalign{\smallskip}
\hline
\hline
\noalign{\smallskip}
HBRNN-L \cite{Du2015Hierarchical}                       &2015             &64.0&59.1\\
Part-aware LSTM \cite{Shahroudy2016NTU}                 &2016             &70.3&62.9\\
Trust Gate ST-LSTM \cite{Liu2016Spatio-temporal}        &2016             &77.7&69.2\\
Two-stream RNN \cite{Wang2017Modeling}                  &2017             &79.5&71.3\\
STA-LSTM \cite{Song2017Attention}                       &2017             &81.2&73.4\\
Ensemble TS-LSTM \cite{Inwoong2017Ensemble}             &2017             &81.3&74.6\\
Visualization CNN \cite{liu2017enhanced}                &2017             &82.6&76.0\\
VA-LSTM \cite{Zhang2017View}                            &2017             &87.6&79.4\\
%MANs \cite{Chunyu2018Memory}                            &2018             &93.2&82.7\\
ST-GCN \cite{Yan2018Spatial}                            &2018             &88.3&81.5\\
SR-TSL \cite{Chenyang2018Skeleton}                      &2018             &92.4&84.8\\
HCN \cite{Chao2018Co-occurrence}                        &2018             &91.1&86.5\\
PB-GCN \cite{Kalpit2018Part}                            &2018             &93.2&87.5\\
\hline
AGC-LSTM (Joint)                                &-   &93.5&87.5\\
AGC-LSTM (Part)                                 &-   &93.8&87.5\\
AGC-LSTM (Joint\&Part)                          &-    &\textbf{95.0}&\textbf{89.2}\\
\hline
\end{tabular}
\end{center}
\caption{Comparison with the state-of-the-art methods on the NTU RGB+D dataset for Cross-View (CS) and Cross-Subject (CV) evaluation in accuracy.}
\label{ntu_result}
\end{table}

From Table \ref{ntu_result}, we can see that our proposed method achieves the best performance of 95.0\% and 89.2\% in terms of two protocols on the NTU dataset. To demonstrate the effectiveness of our method, we choose the following related methods to compare and analyze the results:

\textbf{\emph{AGC-LSTM vs HCN}}.\hspace{1mm}
HCN \cite{Chao2018Co-occurrence} employs the CNN model for learning global co-occurrences from skeleton data. It treats each joint of a skeleton as a channel, then uses the convolution layer to learn the glob co-occurrence features from all joints. We can see that our performances significantly outperform the HCN \cite{Chao2018Co-occurrence} by about 3.9\% and 2.7\% for cross-view evaluation and cross-subject evaluation, respectively.

\textbf{\emph{AGC-LSTM vs GCN models}}.\hspace{1mm}
In order to compare fairly with \cite{Yan2018Spatial}, we use the same GCN operator in the proposed AGC-LSTM layer as in ST-GCN.
%For ST-GCN \cite{Yan2018Spatial}, it applies GCN to model spatial configuration of the joints, then uses the convolutional operator to learn temporal dynamics in each layer.
On the joint-level evaluation, the results of AGC-LSTM are 93.5\% and 87.5\% that outperform 5.2\% and 6.0\% than ST-GCN. Moreover, Our model outperforms the PB-GCN \cite{Kalpit2018Part} by 1.8\% and 1.7\% for tow evaluations. The comparison results prove that the AGC-LSTM is optimal for skeleton-based action recognition than ST-GCN.

%\textbf{\emph{AGC-LSTM vs ST-GCN}}.\hspace{1mm}
%In order to compare fairly with \cite{Yan2018Spatial}, we use the same GCN operator in the proposed AGC-LSTM layer as in ST-GCN. For ST-GCN \cite{Yan2018Spatial}, it applies GCN to model spatial configuration of the joints, then uses the convolutional operator to learn temporal dynamics in each layer. On the joint-level evaluation, the results of AGC-LSTM are 93.5\% and 87.5\% that outperform 5.2\% and 6.0\% than ST-GCN. The comparison results prove that the AGC-LSTM is optimal for skeleton-based action recognition than ST-GCN.

\textbf{\emph{Co-occurrence relationship between spatial and temporal domains}}.\hspace{1mm}
Although Si \etal.\cite{Chenyang2018Skeleton} propose a spatial reasoning and temporal stack learning network with graph neural network (GNN) and LSTM, they ignore the co-occurrence relationship between spatial and temporal domains. Due to the ability to explore the co-occurrence relationship between spatial and temporal domains, Our AGC-LSTM outperforms \cite{Chenyang2018Skeleton} by 2.6\% and 4.4\%.

\textbf{\emph{The performances on joint level and part level}}.\hspace{1mm}
Recent methods can be grouped into two categories: joint-based \cite{Yan2018Spatial, Zhang2017View, Inwoong2017Ensemble, Wang2017Modeling, Chao2018Co-occurrence} and part-based methods \cite{Chenyang2018Skeleton, Wang2017Modeling, Du2015Hierarchical}. Our method achieves the state-of-the-art results on joint-level and part-level, which illustrates the better generalization of our model for joint-level and part-level inputs.

\begin{table}[b]
\begin{center}
\begin{tabular}{lccc}
\hline\noalign{\smallskip}
Methods&Year&Accuracy (\%)\\
\noalign{\smallskip}
\hline
\hline
\noalign{\smallskip}
Lie group \cite{Raviteja2014Human}             &2014        &74.2\\
Actionlet ensemble \cite{Jiang2014Learning}    &2014        &76.0\\
HBRNN-L \cite{Du2015Hierarchical}              &2015        &78.5\\
Visualization CNN \cite{liu2017enhanced}       &2017        &86.1\\
Ensemble TS-LSTM \cite{Inwoong2017Ensemble}    &2017        &89.2\\
\hline
AGC-LSTM (Joint)                               &-        &92.2\\
AGC-LSTM (Part)                                &-        &90.1\\
AGC-LSTM (Joint\&Part)                         &-        &\textbf{93.3}\\
\hline
\end{tabular}
\end{center}
\caption{Comparison with the state-of-the-art methods on the Northwestern-UCLA dataset in accuracy.}
\label{ucla_result}
\end{table}

\subsubsection{Northwestern-UCLA Dataset}

As shown in Table \ref{ucla_result}, the proposed AGC-LSTM again achieves the best accuracy of 93.3\% on the Northwestern-UCLA dataset. The previous state-of-the-art model \cite{Inwoong2017Ensemble} employs multiple Temporal Sliding LSTM (TS-LSTM) to extract short-term, medium-term and long-term temporal dynamics respectively, which has similar functionality to our temporal hierarchical architecture. However, our model outperforms TS-LSTM \cite{Inwoong2017Ensemble} by 4.1\%. Compared with the CNN-based method \cite{liu2017enhanced}, our method also obtains much better performance.

\begin{table}[t]
\begin{center}
\begin{tabular}{c|l|cc}
\hline\noalign{\smallskip}
\multicolumn{2}{c|}{Methods}&CV&CS\\
\noalign{\smallskip}
\hline
\hline
\noalign{\smallskip}
\multirow{4}{*}{Joint}   &LSTM                     &89.4&80.3\\
~                        &GC-LSTM                  &92.4&85.6\\
~                        &LSTM+TH                  &90.4&81.4\\
~                        &GC-LSTM+TH               &92.9&86.3\\
~                        &AGC-LSTM+TH (AGC-LSTM)  &93.5&87.5\\
\hline
Part                     &AGC-LSTM+TH (AGC-LSTM)   &93.8&87.5\\
\hline
\multicolumn{2}{c|}{AGC-LSTM (Joint\&Part)}                    &\textbf{95.0}&\textbf{89.2}\\
\hline
\end{tabular}
\end{center}
\caption{The comparison results between several baselines and our AGC-LSTM on the NTU RGB+D dataset.}
\label{ablation_ntu}
\end{table}

\subsection{Model Analysis}

%To understand the properties of our AGC-LSTM network, we analyze the effectiveness of several key components on both NTU RGB+D dataset and Northwestern-UCLA dataset, \ie temporal hierarchical architecture, AGC-LSTM, attention enhanced mechanism in AGC-LSTM and the two-streams network. Finally, we analyze several failure cases to discuss the existing problems for skeleton-based action recognition.

\subsubsection{Architecture Analysis}

\begin{table}[t]
\begin{center}
\begin{tabular}{c|l|c}
\hline\noalign{\smallskip}
\multicolumn{2}{c|}{Methods}&Accuracy (\%)\\
\noalign{\smallskip}
\hline
\hline
\noalign{\smallskip}
\multirow{4}{*}{Joint}   &LSTM                       &70.0\\
~                        &GC-LSTM                    &87.5\\
~                        &LSTM+TH                    &78.5\\
~                        &GC-LSTM+TH                 &89.4\\
~                        &AGC-LSTM+TH (AGC-LSTM)    &92.2\\
\hline
Part                     &AGC-LSTM+TH (AGC-LSTM)     &90.1\\
\hline
\multicolumn{2}{c|}{AGC-LSTM (Joint\&Part)}                        &\textbf{93.3}\\
\hline
\end{tabular}
\end{center}
\caption{The comparison results between several baselines and our AGC-LSTM on the Northwestern-UCLA dataset.}
\label{ablation_ucla}
\end{table}

Tables \ref{ablation_ntu} and \ref{ablation_ucla} show experimental results of several baselines on the NTU RGB+D dataset and Northwestern-UCLA dataset, respectively. TH denotes temporal hierarchical architecture. Compared with LSTM and GC-LSTM, LSTM+TH and GC-LSTM+TH can increase the temporal receptive fields of each time step on the top layer. The improved performances prove that the temporal hierarchical architecture can boost the ability of representing temporal dynamics. Replacing LSTM with GC-LSTM, GC-LSTM+TH increases the accuracies to 2.5\%, 4.9\% on the NTU dataset and 10.9\% on the Northwestern-UCLA dataset, respectively. Substantial performance improvements verify the effectiveness of GC-LSTM, which can capture more discriminative spatial-temporal features from skeleton data. Compared with GC-LSTM, AGC-LSTM can employ the spatial attention mechanism to select spatial information of key joints, which can promote the ability of feature representation. In addition, the fusion of part-based and joint-based AGC-LSTM can further improve the performance.

\begin{figure}[t]
	\centering
	\subfigure[\fontsize{7}{7}\selectfont ]{
        \label{attention-results:a}
        \includegraphics[width=1.12in]{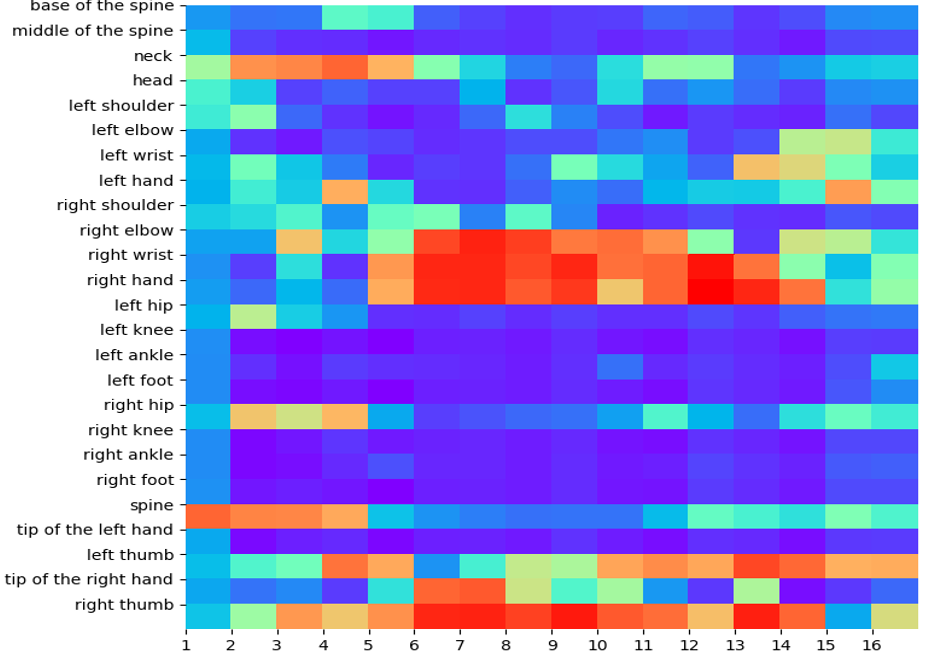}}
    \subfigure[\fontsize{7}{7}\selectfont ]{
        \label{attention-results:b}
        \includegraphics[width=0.90in]{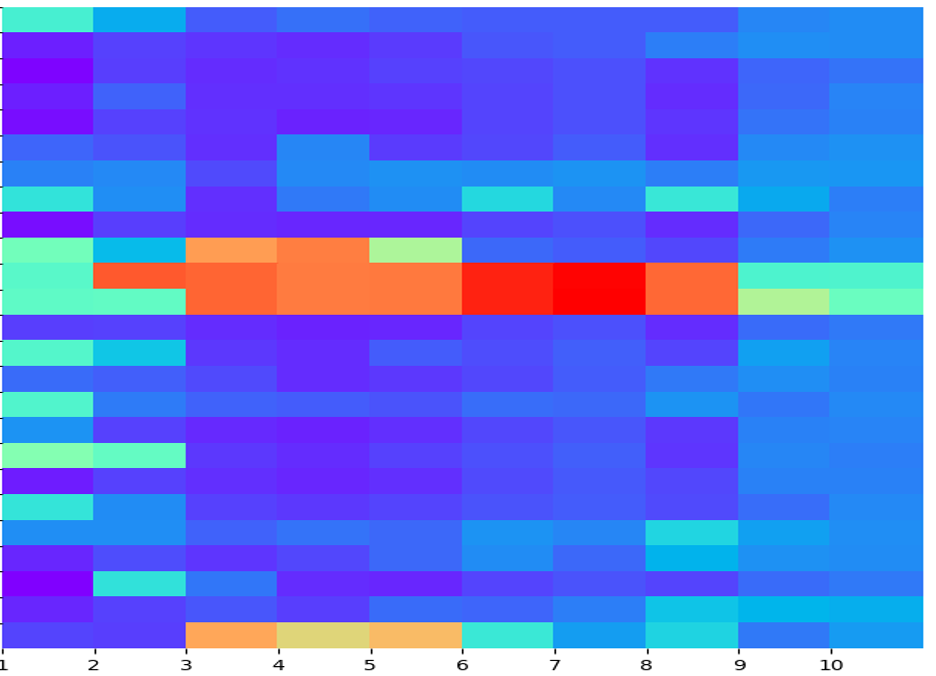}}
    \subfigure[\fontsize{7}{7}\selectfont ]{
        \label{attention-results:c}
        \includegraphics[width=1.07in]{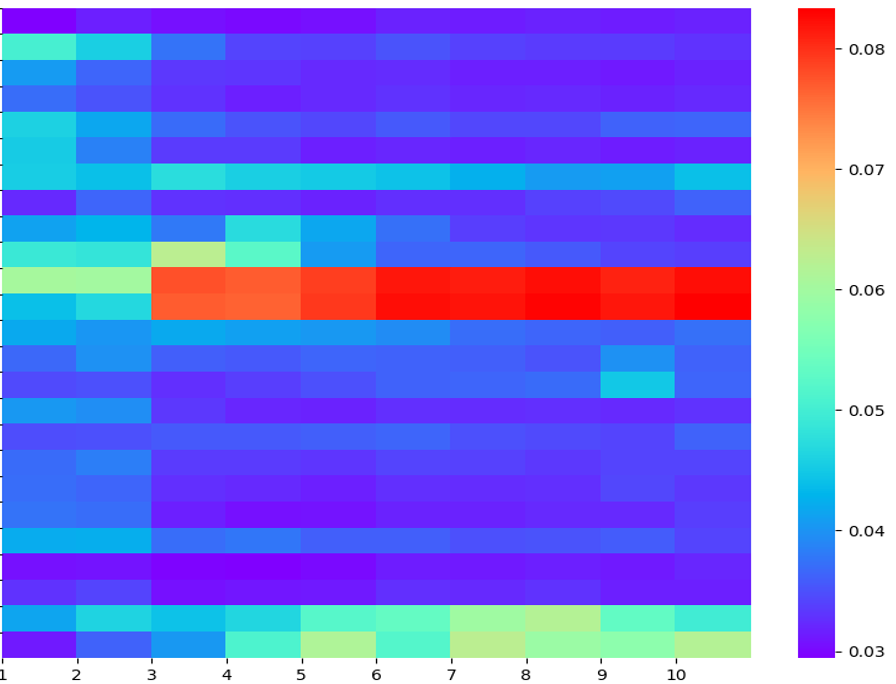}}
   	\caption{Visualizations of the attention weights of three AGC-LSTM layers on one actor of the action ``handshaking``. Vertical axis denotes the joints. Horizontal axis denotes the frames. (a), (b), (c) are the attention results of the first, second and third AGC-LSTM layer, respectively. }
   	\label{attention-results}
\end{figure}

\begin{figure}[b]
	\centering
	\subfigure[\fontsize{7}{7}\selectfont LSTM]{
        \label{NUCLA-acc-results:a}
        \includegraphics[width=1.45in]{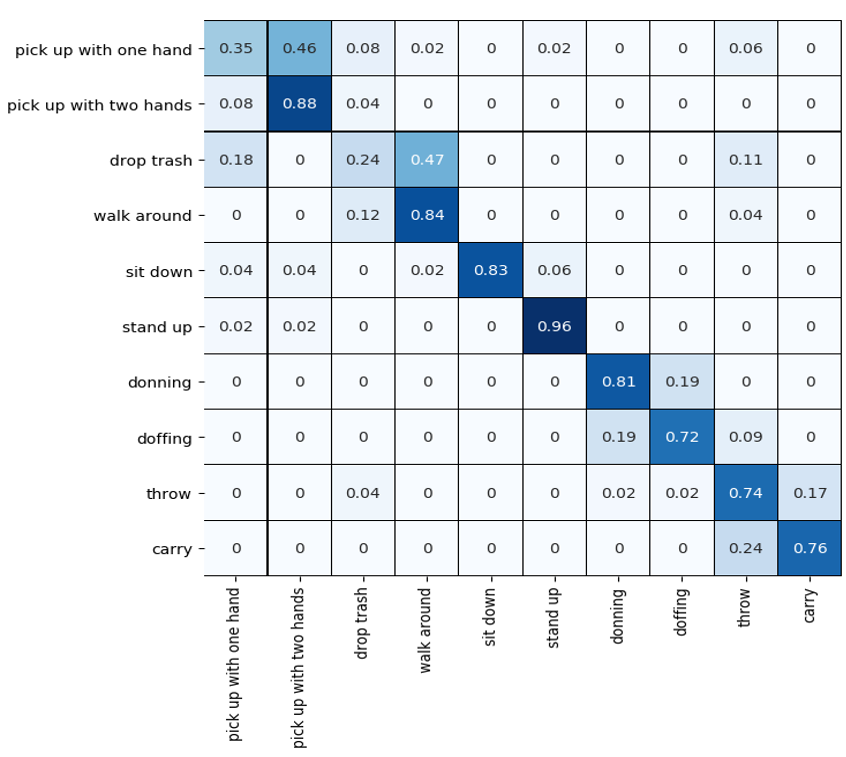}}
    \subfigure[\fontsize{7}{7}\selectfont AGC-LSTM]{
        \label{NUCLA-acc-results:b}
        \includegraphics[width=1.45in]{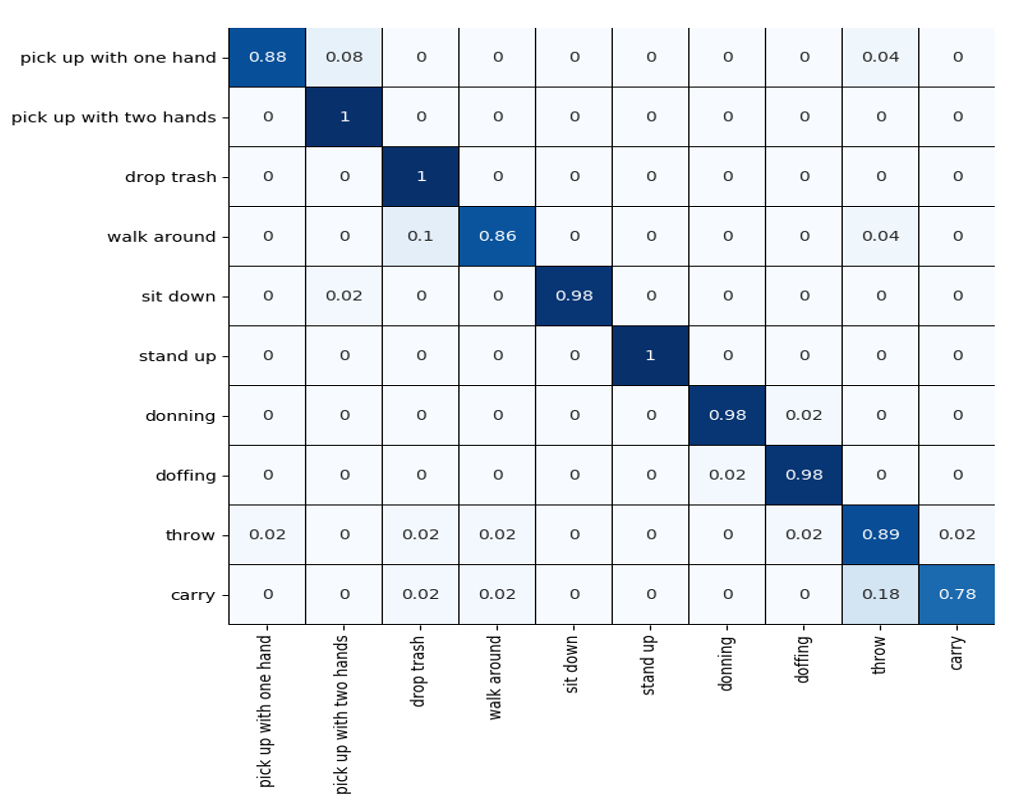}}
   	\caption{Confusion matrix comparison on the Northwestern-UCLA dataset. (a) LSTM. (b) AGC-LSTM.}
   	\label{NUCLA-acc-results}
\end{figure}

We also visualize the attention weights of three AGC-LSTM layers in Fig.\ref{attention-results}. For the ``handshaking'' action, the results show our method can gradually enhance the attention of ``right elbow'', ``right wrist'', and ``right hand''. Meanwhile, ``tip of the right hand'' and ``right thumb'' have some degree of attention.
Furthermore, we analyze the experimental results with a confusion matrix on the Northwestern-UCLA dataset. As show in Fig.\ref{NUCLA-acc-results:a}, it is very confusing for LSTM to recognize similar actions. For example, the actions ``pick up with one hand'' and ``pick up with two hands'' have very similar skeleton sequences. Nevertheless, we can see that the proposed AGC-LSTM can significantly improve the ability to classify these similar actions (shown in Fig.\ref{NUCLA-acc-results:b}). The above results illustrate that the proposed AGC-LSTM is an effective method for skeleton-based action recognition.

\subsubsection{Failure Case}

\begin{figure}[t]
	\centering
	\includegraphics[width=0.9\linewidth]{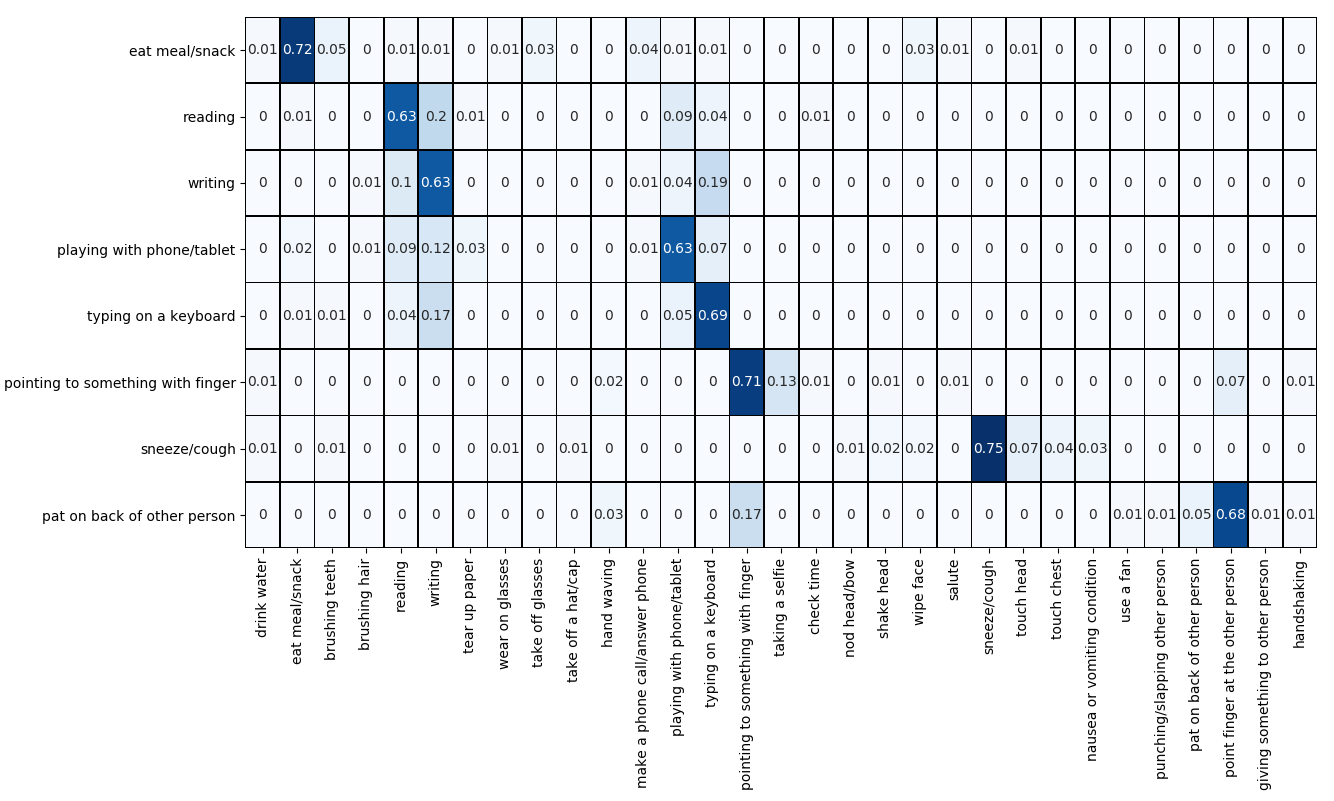}
   	\caption{ Confusion matrix comparison on the NTU dataset.  It shows the part of confusion matrix comparison of the actions (``eat meal/snack'', ``reading'', ``writing'', ``playing with phone/tablet'', ``typing on a keyboard'', ``pointing to something with finger'', ``sneeze/cough'', ``pat on back of other person'') with accuracies less than 80\% on NTU dataset.}
   	\label{NTU-acc}
\end{figure}

Finally, we analyze misclassification results with a confusion matrix on the NTU dataset. Fig.\ref{NTU-acc} shows the part confusion matrix comparison of the actions (``eat meal/snack'', ``reading'', ``writing'', ``playing with phone/tablet'', ``typing on a keyboard'', ``pointing to something with finger'', ``sneeze/cough'', ``pat on back of other person'') with accuracies less than 80\% for the cross-subject setting on the NTU dataset. We can see that misclassified actions are mainly very similar movements. For example, 20\% samples of ``reading'' are misclassified as ``writing'', and there are 19\% sequences of ``writing'' misclassified as ``typing on as keyboard''. For the NTU dataset, only two joints are marked on fingers (``tip of the hand'' and ``thumb''), so that it is very challenging to capture such subtle movements of the hands.

\section{Conclusion and Future Work}

In this paper, we propose an attention enhanced graph convolutional LSTM network (AGC-LSTM) for skeleton-based action recognition, which is the first attempt of graph convolutional LSTM for this task. The proposed AGC-LSTM can not only capture discriminative features in spatial configuration and temporal dynamics, but also explore the co-occurrence relationship between spatial and temporal domains. Furthermore, the attention network is employed to enhance information of key joints in each AGC-LSTM layer. In addition, we also propose a temporal hierarchical architecture to capture high-level spatiotemporal semantic features. On two challenging benchmarks, the proposed AGC-LSTM achieves the state-of-the-art results. Learning the pose-object relation is helpful to overcome the limitations mentioned in the failure case. In the future, we will try the combination of skeleton sequence and object appearance to promote the performance of human action recognition.

\section{Acknowledgements}\label{sect:acknowledgements}

This work is jointly supported by National Key Research and Development Program of China (2016YFB1001000), National Natural Science Foundation of China (61525306, 61633021, 61721004, 61420106015, 61572504), Capital Science and Technology Leading Talent Training Project (Z181100006318030), and Beijing Science and Technology Project (Z181100008918010).

%This work is jointly supported by National Key Research and Development Program of China (2016YFB1001000), National Natural Science Foundation of China (61525306, 61633021, 61721004, 61420106015, 61572504).

{\small
\bibliographystyle{ieee_fullname}
\bibliography{egbib}
}

\end{document}